\documentclass{article}

\PassOptionsToPackage{numbers, sort&compress}{natbib}
\usepackage[final]{neurips_2020}

\usepackage[utf8]{inputenc} 
\usepackage[T1]{fontenc}    
\usepackage{hyperref}       
\usepackage{url}            
\usepackage{booktabs}       
\usepackage{amsfonts}       
\usepackage{nicefrac}       
\usepackage{microtype}      

\usepackage{amsmath,graphicx}
\usepackage{subcaption,siunitx,booktabs}
\usepackage{multirow}
\usepackage{makecell}
\usepackage[table, xcdraw]{xcolor}

\usepackage{colortbl}
\definecolor{mygray}{gray}{.9}

\title{CheerBots: Chatbots toward Empathy and Emotion using Reinforcement Learning}

\author{
    Jiun-Hao Jhan \quad Chao-Peng Liu \quad Shyh-Kang Jeng \quad Hung-Yi Lee \\
    National Taiwan University, Taipei, Taiwan \\
    \texttt{\{a25200035, liu0958101584\}@gmail.com \quad \{skjeng, hungyilee\}@ntu.edu.tw}\\
}

\begin{document}

\maketitle

\begin{abstract}
  Apart from the coherence and fluency of responses, an empathetic chatbot emphasizes more on people's feelings. By considering altruistic behaviors between human interaction, empathetic chatbots enable people to get a better interactive and supportive experience. This study presents a framework whereby several empathetic chatbots are based on understanding users' implied feelings and replying empathetically for multiple dialogue turns. We call these chatbots CheerBots. CheerBots can be retrieval-based or generative-based and were finetuned by deep reinforcement learning. To respond in an empathetic way, we develop a simulating agent, a Conceptual Human Model, as aids for CheerBots in training with considerations on changes in user's emotional states in the future to arouse sympathy. Finally, automatic metrics and human rating results demonstrate that CheerBots outperform other baseline chatbots and achieves reciprocal altruism. The code and the pre-trained models will be made available.
\end{abstract}

\section{Introduction}
It is empathy that is key for a chatbot to chat with users in a more human-like way. In addition to the coherence of chatbots' responses, people might place more emphasis on their feeling during chatting with chatbots. That is, responding to a user by understanding any implied feelings is a desirable attribute in developing a dialogue system. We regard this kind of dialogue systems as empathetic chatbots. Some researchers claimed that empathy involves being tenderhearted toward another person \citep{Snyder_Positive}. By considering the feelings of others, people can have a better interactive and supportive experience. For instance, given “I failed to pass the exam yesterday” as the speaker’s opening sentence, both responses “Again? You should have passed!” and “Don’t worry. You will pass next time.” are relevant and acceptable responses. Considering the emotional states, however, the latter response would make the speaker feel better, whereas the former response seems to blame the first speaker, making the speaker feel worse than before. The latter response may be more satisfactory since it acknowledges the speaker's underlying feelings in an empathetic manner.

To incorporate empathy into a chatbot, developers concentrate on humans' emotional states. \citet{lin2019caire} extended a learning approach \citep{wolf2019transfertransfo} on an empathetic dialogue dataset and finetuned their proposed chatbot with extra emotion classification objectives to produce more emotion-evoking responses. However, we argue that current empathetic chatbots suffer from a limited ability to arouse sympathy since they overlook the altruistic behavior between human interaction. In general, empathy is regarded as an ability to respond emotionally based on other people's emotional or arousal state \citep{Waal2008PAB}. It enables a human to learn more about the other's psychological feelings and give more emotional support. Thus, an empathetic chatbot should consider both the speaker's and the listener's roles in a conversation: changes between these two interlocutors' emotional states after conversing should be considered.

In this paper, we propose an innovative framework whereby chatbots are trained to achieve reciprocal altruism. These chatbots, including retrieval-based and generative-based, are named CheerBots. We will examine the difference between the interlocutors' emotional states as empathy valence. In this regard, CheerBots generate a response to amplify the empathy valence, given inputs, and the predicted emotion. The emotion that arouses empathy valence is predicted via our finetuned emotion predictor using reinforcement learning (RL) \citep{williams1992rl}. Consequently,  empirical results reveal that CheerBots outperform existing chatbots on empathetic response generation. The contributions of this paper hence can be 2-fold:
\begin{itemize}
\itemsep=-3pt
    \item [1)]
    In our proposed framework, to amplify user empathy, we establish an agent simulating a human (Conceptual Humans Model) to consider possible changes between the interlocutors' future emotional states.
    \item [2)]
    The experimental results demonstrate that CheerBots achieve state-of-the-art performance on the EmpatheticDialogues (ED) dataset \citep{rashkin-etal-2019-towards}.
\end{itemize}

\section{Related Work}
Several researchers applied deep neural networks to build coherent chatbots by retrieval-based \citep{zhou-etal-2016-multi, wu-etal-2019-sequential, yang2018learning, henderson2017efficient, Yan2016LRDNN} and generative-based \citep{ritter-etal-2011-data, Serban2016BE2E, shang-etal-2015-neural, Tammewar2017ProductionRC} approaches. Even though these chatbots have been intelligent enough to converse fluently, they still cannot well imitate dialogues between human beings. It is mainly because they are essentially emotionless. To endow emotion to a chatbot, several works proposed to control the generated sentence with a given sentiment. They replaced the attribute phrases with phrases of the target attribute \citep{feng-etal-2018-pathologies, li2016understanding, li-etal-2018-delete}. However, they are still possible in wrong phrases since they can not really understand the context's meanings and might damage the sentence's overall consistency. Some researchers manage to generate the target style sentence, transfer the sentence's style, and add emotional tags to express sentences' emotions \citep{hu2017controlled, wang2019controllable, niu2018polite}.

Nonetheless, a more human-like chatbot requires the ability to realize others' underlying feelings. To choose the appropriate response, developers should emphasize more on empathy. Several works proposed data-driven approaches to predict the current emotion and generate a response jointly. \citet{rashkin-etal-2019-towards} proposed a new benchmark for empathetic dialogue generation, the ED dataset, grounded in emotional situations. Based on the ED dataset, \citet{lin2019caire} extended the open domain chit-chat models with a dialogue-based emotional loss to enhance the model's ability to express emotion. Moreover, to explore the user's feelings after chatting with the chatbot, \citet{shin2019happybot} used deep RL incorporated with a user-sentiment approximator to improve the performance of a chatbot. Their method is grounded in the two-virtual-agents architecture \citep{DBLP:journals/corr/LiMRGGJ16} to predict the possible responding emotion.

\begin{figure*}
\begin{center}
\includegraphics[width=1.0\linewidth]{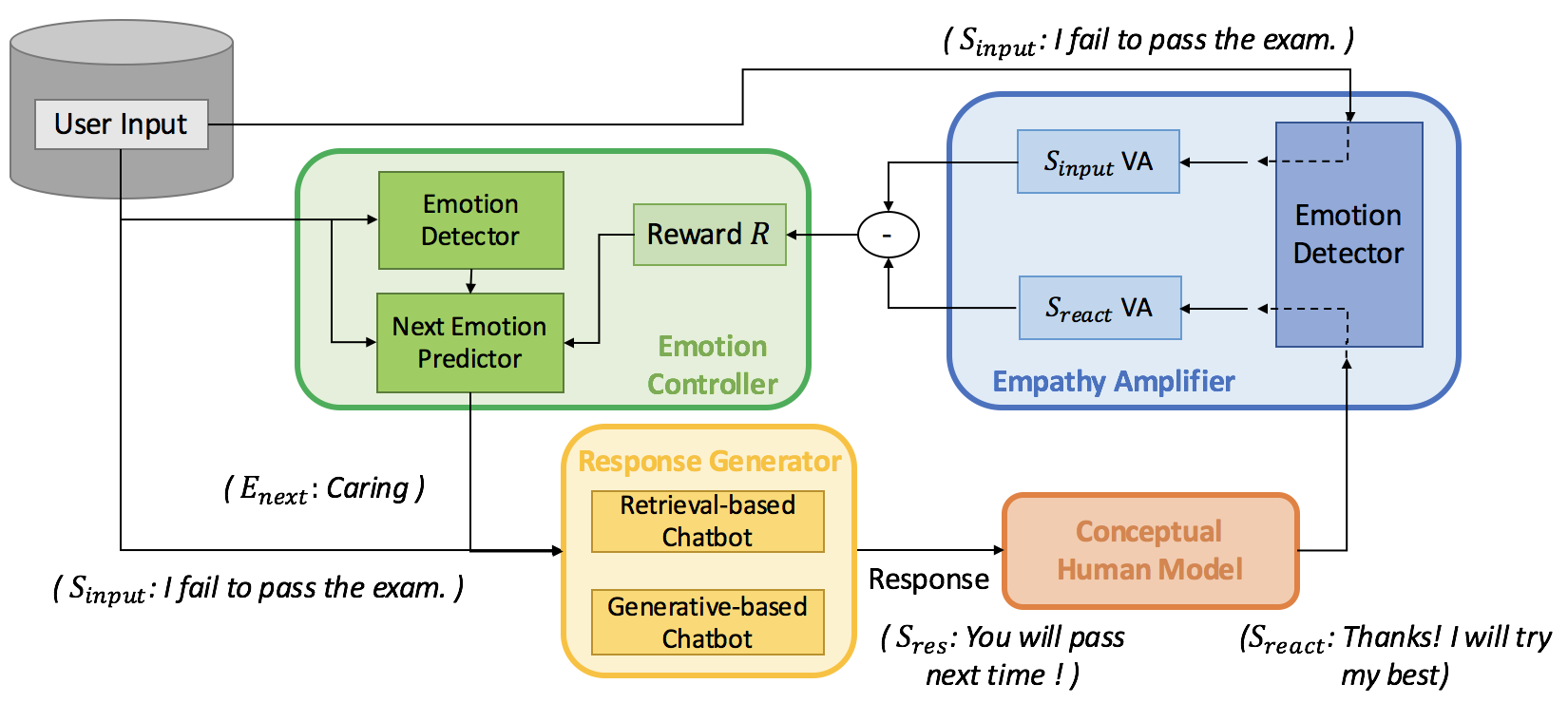}
\vspace{-4mm}
\end{center}
   \caption{Framework overview. The Emotion Controller is in charge of detecting and predicting the next emotion; the Response Generator is responsible for conditionally generating responses;  the Empathy Amplifier aims to optimize the empathy valence. The utterances in the brackets serve as examples.}
   \label{fig:overview}
\label{fig:framework}
\end{figure*}

However, different from previous works, CheerBots directly learn from the speaker's feedback instead of probability, which might engender variance. Theoretically, sentences allow us to self-evaluate models, whereas estimated probabilities of emotions imply nothing when measuring \citep{shin2019happybot}. Technically, ours is more flexible to leverage data against an end-to-end model. Besides, we extended two-virtual-agents architecture to the speaker-listener architecture, assigning different objectives to each agent. CheerBots, acting as a listener, is aimed to be empathetic, while the Conceptual Human Model, acting as a speaker, is aimed to simulate humans generally. As a result, CheerBots are able to learn how to respond empathetically from the interaction between interlocutors.

\section{Empathetic Chatbot}
\subsection{Framework Overview}
Figure \ref{fig:overview} displays the entire framework pipeline composed of four major components: Emotion Controller, Response Generator, Conceptual Human Model (CHM), and Empathy Amplifier. First of all, the Emotion Controller will detect the emotion $E_{now}$ of the input sequence $S_{input}$ and predict the next emotion $E_{next}$ to make the chatbot generate a more empathetic response. Hence, the Response Generator should take both $S_{input}$ and $E_{next}$ into considerations to produce a response $S_{res}$. In light of $S_{res}$, CHM, which simulates a human involved in the conversation, will give a corresponding reply $S_{react}$, namely a probable response the speaker will react. In the end, the Empathy Amplifier will gauge the emotional difference between $S_{input}$ and $S_{react}$ as empathy valence $R$. In this regard, we can maximize the empathy valence to finetune the Next Emotion Predictor using deep RL so as to make the chatbot arouse sympathy.

\subsection{Emotion Controller}
\label{sec:ec}
The Emotion Controller is responsible for all emotion-related processes and contains two units: Emotion Detector and Next Emotion Predictor.

\paragraph{Emotion Detector}
\label{sec:ed}
We employ a Valence-Arousal (VA) projection based on the Valence-Arousal coordinate \citep{model_of_affect, VA_guidance}. Each emotion can be represented as two-dimensional vectors\footnote{e.g. afraid=[-0.12, 0.79], joyful=[0.85, 0.15].}. Emotion Detector, trained by BERT \citep{devlin2018bert} for improving language understanding, is used to detect the emotion and its corresponding VA of the current sentence $S_{input}$. This unit uses two loss functions composed of the cross-entropy loss $L_d$ and L2 norm loss with different weights $L_c$ to optimize the network. $L_d$ represents the cross-entropy loss for classification, and $L_c$ represents the VA coordinates for the predictions. Given input sequence $x_i$ and Ground Truth $g_i$, we finetuned the model on the emotion classification task to enable the model to categorize emotions from the input sequence by the cross-entropy loss: $L_d$ loss sums the negative log-likelihood for all the training sentences and penalizes each class's classification error equally.
Then, to avoid the model from being confused by similar emotions ( e.g., furious and angry), we provide VA values in the training process. Specifically, we add the Valence-Arousal projection layers after the classification layer to compute an L2 norm loss $L_C$. 
\begin{equation}
\label{eq:ED_l2norm}
    L_c = \frac{1}{N} \sum_{i=1}^{N} {(V_{x_i} - V_{g_i})}^2 + {(A_{x_i} - A_{g_i})}^2   
\end{equation}
$V, A$ are the projected valence and arousal values according to the detected emotion of sequence $x_i$ or Ground Truth emotion $g_i$. With the sharing of features, the model can distinguish relatively similar emotions. As a result, the label with the maximum probability in the classification output layer is treated as the dominant emotion label. 

\paragraph{Next Emotion Predictor}
The Next Emotion Predictor is aimed to predict the responding emotion for Response Generator, and we apply the same BERT architecture to train. Given the input sentence $S_{input}$ and its detected emotion $E_{now}$, we concatenate the semantic vectors embedded by the BERT encoder with the one-hot emotion class to form a vector with a more meaningful understanding of the sentence. Then concatenated vectors are passed to a linear classifier to predict the probability of the next emotions. Also, the same cross-entropy loss mentioned in Section \ref{sec:ed} is applied for emotion prediction loss. Therefore, Response Generator can respond with an appropriate emotion to arouse sympathy. We will further illustrate the finetuning processes of adjusting the Next Emotion predictor to augment the empathy valence in Section \ref{sec:emp_amp}.

\subsection{Response Generator} 
Response Generator aims to produce a response according to the given predicted emotion. We implemented retrieval-based and generative-based chatbots in this unit.

\paragraph{Retrieval-based Chatbot}
In the retrieval-based setup, we adopt the previous method \citep{rashkin-etal-2019-towards} to train a retrieval-based model. Given the context $X$, the model aims to choose the best response in candidates group $Y$. In training, we apply the BERT encoder to encode both contexts and candidates: the context $x \in X$ is tokenized into $x_1, x_2, ...$, encoded into $h_x$; meanwhile, the candidate $y \in Y$ is tokenized into $y_1, y_2, ...$, encoded into $h_y$. Hence, we employ a softmax function to pick up an appropriate response among candidates $y_i$, and, therefore, the objective function that minimizes the negative log-likelihood to choose the best candidate $y^\star$ is denoted as below:
\begin{equation} 
\label{eq:ret}
    y^\star = - argmax( h_x \cdot h_y)
\end{equation}
To set up negative samples in training, we used other candidates in the same batch as negative samples. Moreover, to produce responses with a particular emotion, we implemented an \textbf{Emotion Filter} to retrieve candidates within given emotions. All candidates were divided into several groups according to the emotions mentioned in Section \ref{sec:ec}. The Emotion Filter will only retrieve candidates from a particular emotion group corresponding to the predicted emotion selected by the Next Emotion Predictor.

\paragraph{Generative-based Chatbot}
\label{sec:gen_chat}
In the generative model, we apply the OpenAI GPT architecture \citep{Radford2018ImprovingLU}. Firstly, we improve the consistency and engagement of the model by adding an additional segment $S_P$ to the commonplace generative chatbot format \citep{lin2019caire}: we concatenate customized persona (e.g., “I like to help people.") at the start of each input, including the dialogue history segment $S_H$ and the Ground Truth reply $S_R$. With this customized persona, our chatbot is enhanced to care for others while responding. The training loss function composed of the language model (LM), next-sentence prediction (NSP), and emotional sentence generation (ESG) loss, which is defined as eq (\ref{eq:gen})
\begin{equation} 
\label{eq:gen}
    L = L_{LM} + L_{NSP} + L_{ESG}  
\end{equation}
$L_{LM}$ is computed by cross-entropy loss in language model training. As for $L_{NSP}$, we randomly sample distracted sentences from the dataset as distractors and train the model encoder to distinguish whether an input sequence ends with a correct reply or a distractor. It leads the model to look at the global segments as well as the local context via the binary cross-entropy. Furthermore, to generate a response with a particular emotion, we extend the idea proposed by \citet{niu2018polite}. To elaborate, we added vocabulary emotion labels with trainable word embeddings. Then, we prepend the emotion label in the input sequence and add an emotion classification loss $L_{ESG}$ to ensure the encoded vector can conform to the prepended label using cross-entropy. As a result, the input format consists of three segments and an emotional label, and we can predict the target sequence \^{y} through three different loss functions.

\begin{table*}
\centering
\caption{Performance of quantity experiment results. \textbf{Retrieval}: (1) Baseline: pre-trained model from \citet{rashkin-etal-2019-towards} (2) Finetuned: our model reproduced based on \citet{rashkin-etal-2019-towards} with BERT model. (3) Finetuned-EF: a finetuned model with emotion filter to retrieve candidates only from predicted emotion. (4), (5) PG-Finetuned and DQN-Finetuned: models considering the change of emotion valence from the speaker finetuned by the deep RL. \textbf{Generative}: (6) Baseline: finetuned model referred from \citet{rashkin-etal-2019-towards}. (7) CAiRE: generative model proposed by \citet{lin2019caire}. (8) Finetuned: model finetuned on the ED dataset. (9) EmoPrepend: model trained with emotion prepending and emotion classification loss. (10),  (11) PG-Prepend and DQN-Prepend: models finetuned through deep RL. Higher is better for $P@1,100$ and AVG BLEU. PPL (perplexity) is lower, the better.}
\vspace{3mm}
\begin{tabular}{clccc}
\toprule{}
 &  \textbf{Model} &  \textbf{P@1,100} & \textbf{ PPL} & \textbf{AVG BLEU}\\
\specialrule{0.05em}{3pt}{3pt}
\multirow{5}{*}{\textbf{Retrieval}} & (1) Baseline \citep{rashkin-etal-2019-towards} & 54.13 & - & 5.35 \\
 & (2) Finetuned & 62.23 & - & 5.86  \\
 & (3) Finetuned-EF & 65.08 & - & 5.98 \\
 & (4) PG-Finetuned & 21.13 & - & 5.93  \\
 & (5) DQN-Finetuned & 24.75 & - & 6.11   \\
\specialrule{0.05em}{3pt}{3pt}
\multirow{5}{*}{\textbf{Generative}} & (6) Baseline \citep{rashkin-etal-2019-towards} & - & 21.24 & 6.27 \\
 & (7) CAiRE \citep{lin2019caire} & - & 13.32 & 7.03  \\
 & (8) Finetuned & - & 13.12 & 7.54  \\
 & (9) EmoPrepend & - & 13.34 & 8.02 \\
 & (10) PG-Prepend & - & - & 8.06  \\
 & (11) DQN-Prepend & - & - & 7.9  \\
\bottomrule{}
\end{tabular}
\label{tab:auto}
\end{table*}

\subsection{Conceptual Human Model}  
We created a Conceptual Human Model (CHM) to simulate human behaviors. This model serves as a conceptual agent whereby we can obtain a possible speaker's future response $S_{react}$. A conceptual model, including behavior and interaction, can be regarded as an agent, which describes the semantics of implementation at a higher level of abstraction \citep {Conceptual_Modeling}. Here we use a generative-based chatbot to simulate the speaker's behavior. To establish this conceptual model, we adopted the same training approach described in Section \ref{sec:gen_chat}. The major difference between this model and the generative-based chatbot is the input format for training. The ED dataset participants can be divided into two different parts: the speaker and the listener. The chatbot in the Response Generator used the listener's sentences as training replies, while the CHM used the speaker's sentences as training responses. Also, in the persona segment of the input format, the generative-based chatbot applied the self-defined persona. In contrast, the conceptual model adopted the "situation prompt" of the speaker provided in the ED dataset as the conversation situation. This alternation allows the CHM to generate the responses confined to the given situation.

\subsection{Empathy Amplifier}
\label{sec:emp_amp}
\citet{Waal2008PAB} emphasizes that an empathetic chatbot should apprehend more about each other's underlying feelings and give more emotional support in return: we ought to consider the altruistic behavior between human interaction to learn the underlying feeling. Along this line, we conclude that the altruistic behavior indicates mutual interaction between interlocutors and that underlying feelings imply changes between interlocutors' emotional states. In this regard, CheerBots can examine the difference between the speaker's current emotional states $V(S_{input})$ and the subsequent emotional states $V(S_{react})$ from the CHM chatbot. These emotional differences are represented as empathy valence $R$. Since the speaker will feel sympathy as the empathy valence is positive, we can amplify the empathy valence $R$ using deep RL.
\begin{equation}
\label{eq:reward}
R = V_{S^n_{react}} - V_{S_{input}}
\end{equation}
Moreover, we considered the multi-turns conversation to examine empathetic changes among the interlocutors' different emotional states. For multi-turn conversations, we can obtain several emotional states from the speaker, denoted as $V_{S_{input}}, V_{S^1_{react}}, V_{S^2_{react}}, ..., V_{S^n_{react}}$ where $n$ is the number of turns. Therefore, we can acquire the final empathy valence in multi-turn conversations by aggregating each empathy valence in each turn, which means the difference between the first and final sequences. We utilize these differences as the final reward to finetune the Next Emotion Predictor. The reward function is shown as eq (\ref{eq:reward}). Here we employed the deep RL algorithm with policy gradient (PG) and deep Q-learning (DQN).

\section{Experimental Setup} 
\subsection{Dataset}
\label{sec:dataset}
\paragraph{EmpatheticDialogues Dataset}
\citet{rashkin-etal-2019-towards} established a novel dataset, the EmpatheticDialogues dataset \footnote{https://github.com/facebookresearch/EmpatheticDialogues}, with around 25K communications. The dataset includes 32 different emotion labels covering a wide range of positive, neutral, and negative emotions. Creating a situation related to specific emotional experiences is the goal of offering a single emotion label. They make a great effort to ensure balanced emotion coverage by asking the participants to select the emotion labels evenly. However, some emotions are quite closely relevant in the ED dataset. Thus, we merge the “prepared” emotion into the “confident” emotion, the “sentimental” emotion into the “nostalgic” emotion, and the “terrified” emotion into the “afraid” emotion. We expect that CheerBots can fit the empathetic distribution in the dataset by finetuning.

\subsection{Model Training}

\paragraph{Valence-Arousal Coordinate Projection}
Since \citet{VA_guidance, model_of_affect} only evaluated 19 emotions' values, but the ED dataset contains 32 different emotions, we first supervised train the Emotion Detector grounded in these known emotions. Next, we obtain remained emotions by predicting the pseudo labels from this Emotion Detector. Therefore, we can train another Emotion Detector to map emotions on the VA Coordinate.  

\paragraph{Training Detail}
We load our pre-trained Emotion Detector, Next Emotion Predictor, Response Generator, and the CHM in advance during the deep RL training procedure. Only the Next Emotion Predictor is trainable, whereas other models' weights are fixed. Response Generator can be either a retrieval model or a generative model. The generative model is the only one that finetunes on the PersonaChat \citep{zhang-etal-2018-personalizing} and then finetunes on the ED dataset. Other models are all finetuned on the ED dataset directly. The details of training hyperparameters and infrastructure settings are described in Appendix \ref{sec:hyperparameters} and \ref{sec:spe_details}. 

\subsection{Model Evaluation Setup}
We evaluate the performance of CheerBots against other baselines in two experiments: automatic metrics and human ratings.

\paragraph{Automatic Metrics}
To automatically compare CheerBots with others, we compute the sentence level BLEU score \citep{papineni2002BLEU} for the model response generated by the Response Generator. The computed scores were compared against the actual response in the ED dataset. We also report the perplexity for the generative-based chatbot and calculate $P@1,100$ in the retrieval-based chatbot. $P@1,100$ means the accuracy of retrieving the correct candidate out of 100 candidates. When computing $P@1,100$, the correct response is contained in the candidates, unlike inference from the retrieval models for all other metrics. 

\paragraph{Human Ratings}
In the ratings, participants were given various conversations. Each conversation contains three sentences: a speaker’s opening statement, a corresponding response, and a reply based on the previous response. These three sentences can be referred to $S_{input}$, $S_{res}$, and $S_{react}$, respectively. We design two tasks for participants: Independent Comparison and Pairwise Comparison. (1) For each response in the Independent Comparison, the participants were asked to score from 1 to 5 based on relevance and fluency introduced by \citep{rashkin-etal-2019-towards}. Also, they need to score the increase of the speaker's empathy level according to the whole conversation. (2) As for Pairwise Comparison, we set up several matchups that include two models' different responses according to the same $S_{input}$. Participants need to pick up a model's response that shows more empathy. Further details are shown in Appendix \ref{sec:human_details}.

\begin{table}[t!]
 \begin{minipage}[c]{0.46\textwidth}
 \caption{Deep Reinforcement Learning Results. A three-turn dialogue contained three conversation turns, where a conversation turn includes a speaker and a listener. Similarly, a one-turn conversation contains only one conversation turn. Rewards in three-turn conversations are all greater than those in one-turn conversations.}
  \label{tab:rl}
  \vspace{3mm}
  \begin{tabular}{cccc}
    \toprule{}
    \multirow{2}{*}{\textbf{Model}} & \multirow{2}{*}{\textbf{RL}} & \textbf{Rewards} & \textbf{Rewards}\\
    & & (3 turns) & (1 turn) \\
    \specialrule{0.05em}{3pt}{3pt}
    \textbf{Baseline} & - & 0.13 & -0.002 \\
    \specialrule{0.05em}{3pt}{3pt}
    \multirow{2}{*}{\textbf{Retrieval}} & PG & 0.202 & 0.084 \\
     & DQN & \textbf{0.317} & \textbf{0.11} \\
    \specialrule{0.05em}{3pt}{3pt}
    \multirow{2}{*}{\textbf{Generative}} & PG & 0.222 & 0.07 \\
     & DQN & 0.157 & 0.04 \\
    \bottomrule{}
  \end{tabular}
 \end{minipage}
 \hspace{7mm}
 \begin{minipage}[c]{0.48\textwidth}
  \centering
  \includegraphics[width=\columnwidth]{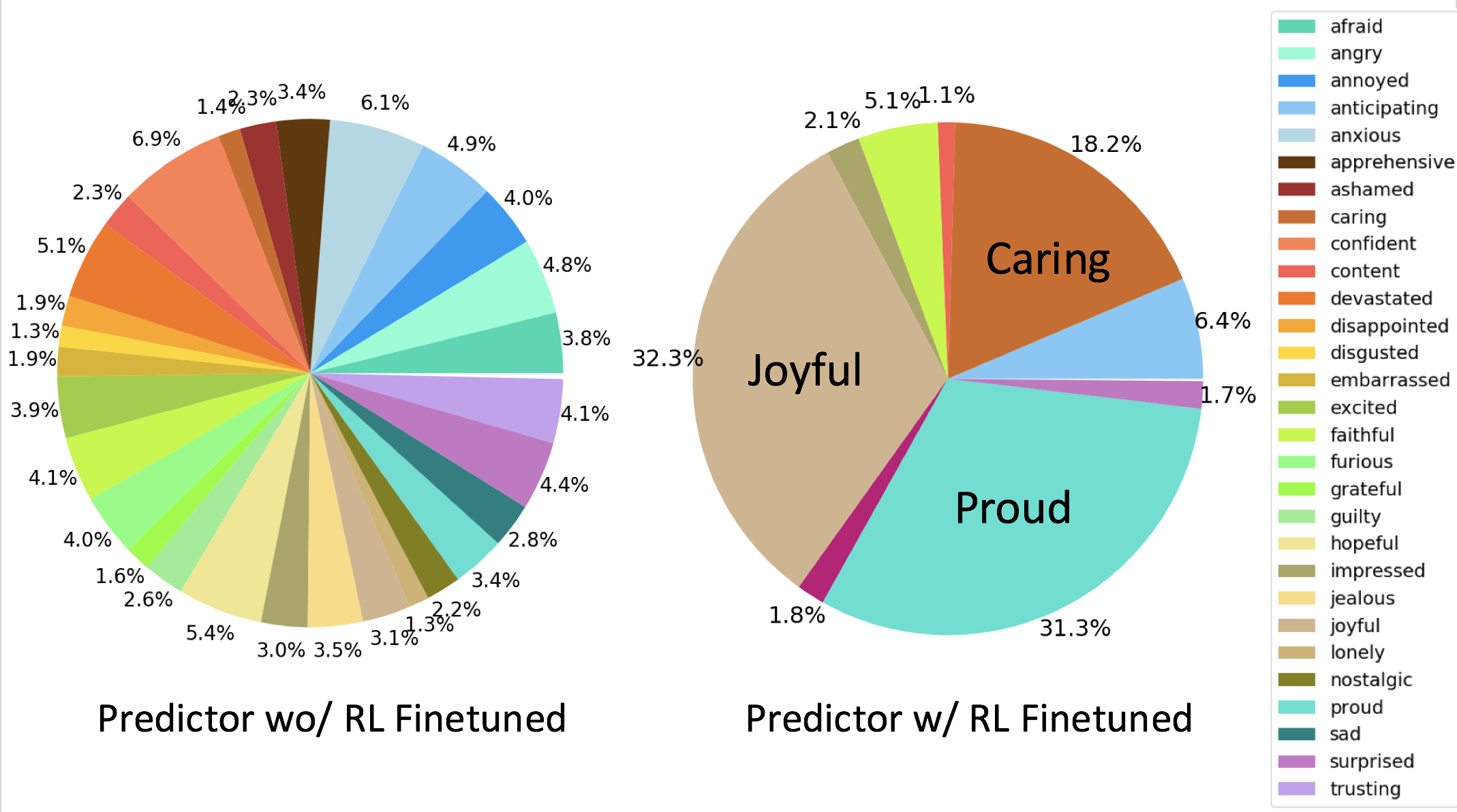}
  \vspace{3mm}
  \captionof{figure}{This figure demonstrates the distributions of the predicted emotion in multi-turns conversation after being finetuned by deep RL.}
  \label{fig:predicted}
 \end{minipage}
\end{table}

\begin{table}[]
\caption{Human Ratings: Pairwise comparison. The setups of each model are the same as Table \ref{tab:auto}. GT (Ground Truth) is a human example sampled from the ED dataset. "Left" means the proportion of participants choose the model on the left side of the model pair, while the "Right" means the opposite. "Tie" means participants cannot choose either side. Top results are boldfaced.}
  \label{tab:pair}
  \vspace{3mm}
  \centering
  \begin{tabular}{cccccc}
    \specialrule{0.05em}{5pt}{5pt}
    \multicolumn{3}{c}{\textbf{Matchup}} & \textbf{Left} & \textbf{Tie} & \textbf{Right} \\ 
    \specialrule{0.05em}{3pt}{3pt}
    (1) Ret-baseline \citep{rashkin-etal-2019-towards} & - & (3) Ret-Finetuned-EF & 52.6\% & 16.3\% & 31.1\% \\
    (1) Ret-baseline \citep{rashkin-etal-2019-towards} & - & (5) Ret-DQN-Finetuned & 31.3\% & 18.2\% & \textbf{50.5\%} \\
    (3) Ret-Finetuned-EF & - & (5) Ret-DQN-Finetuned  & 28.3\% & 19\% & \textbf{52.7\%} \\ 
    \specialrule{0.05em}{3pt}{3pt}
    (8) Gen-Finetuned & - & (10) Gen-PG-Prepend & 30.3\% & 23.5\% & \textbf{46.2\%} \\ 
    \specialrule{0.05em}{3pt}{3pt}
    (5) Ret-DQN-Finetuned & - & (10) Gen-PG-Prepend & \textbf{45.1\%} & 24.3\% & 30.6\% \\
    (5) Ret-DQN-Finetuned & - & Ground Truth & 48.1\% & 18.7\% & 33.2\% \\
    (10) Gen-PG-Prepend & - & Ground Truth & 36.3\% & 17.5\% & 46.2\% \\ 
    \bottomrule{}
   \end{tabular}
\end{table}

\section{Result and Discussion}
\subsection{Automatic Metric Result}
Table \ref{tab:auto} shows the performance of our proposed models in automatic metrics. In retrieval models, by using the ED dataset as candidate sources, the emotion filter increases $P@1,100$ by 3\%. However, in RL-based finetuned retrieval models, the scores in $P@1,100$ are relatively low, but the Avg BLEU scores are higher than the baseline. One of the reasons is that deep RL emphasizes more on rewards. The Next Emotion Predictor prefers the emotion that augments the empathy valence rather than the emotion relevant to the conversation. Therefore, the finetuned Next Emotion Predictor will predict an emotion different from the Ground Truth, making the Ground Truth emotion exclude from the retrieval pool. Nonetheless, the avg BLEU scores of our models still surpass others. Apparently, this result implies that our models can retrieve a response similar to the Ground Truth and arouse sympathy.

In generative-based models, our models outperform the baseline model on both perplexity and Avg BLEU. This means our model provided more accurate sentences that conform to the correct grammar. However, in emotions prepending and RL-based finetuned models, the perplexity scores are relatively worse, but the corresponding Avg BLEU scores are higher than the finetuned model. The reason might be that RL-based learning approaches put more emphasis on empathy. Also, since the Ground Truth sentence is collected in an empathetic manner, our RL-based model can generate sentences more similar to the Ground Truth sentence. Hence, the automatic evaluation results demonstrated that our proposed models could improve the responses' coherence in a conversation.

\subsection{Deep Reinforcement Learning Results}
We demonstrate the performance of deep RL on the test set of the ED dataset. In Table \ref{tab:rl}, the baseline model represents the supervised model, finetuned only on the ED dataset. We examine the average rewards in conversations with three turns and one turn, respectively. We compute the rewards from the speaker’s emotional states between the first and the last turn during the conversation, and we can observe that rewards in three turns are larger than those in only one turn. This illustrates that our models can help the speaker turn to a positive emotional state. According to Figure \ref{fig:predicted}, the results of our emotion predictor meet our expectations. Initially, the predicted emotions are distributed uniformly since we only train the Emotion Predictor on the ED dataset. After finetuned by deep RL, our emotion predictor is capable of selecting some particular emotions to optimize the empathy valence, such as joyful, proud, and caring. In this regard, we can get the largest empathy valence by selecting these kinds of emotions.

Furthermore, as shown in Table \ref{tab:rl}, in the retrieval-based model, the performance of DQN outperforms that of PG. Because DQN needs to examine each action, DQN is good at training with discrete actions. We set up the environment as a dialogue system, and the actions of choosing responding emotions are discrete. As a result, the DQN algorithm can perform relatively well. As for the generative-based models, they are generally worse than the retrieval-based models in both RL methods. It is mainly because bias contained in the emotional sentence generation brings about fluctuation during the RL training. That is, the conditional generator might make mistakes, as the generator might not generate the sentence with entirely correct emotion even though we achieve SOTA performance. Hence, deep RL methods can hardly learn stable policy to select the correct emotion to augment rewards. 

\begin{table}[t]
\centering
\caption{Human ratings: Independent Comparison. The setups of each model are the same as Table \ref{tab:auto}. GT (Ground Truth) is a human example sampled from the ED dataset. Top results are boldfaced.}
\vspace{3mm}
\begin{tabular}{clccc}
\toprule{}
  & \textbf{Model} & \textbf{Empathy} & \textbf{Coherence} & \textbf{Fluency} \\ \specialrule{0.05em}{3pt}{3pt}
\multirow{3}{*}{\textbf{Retrieval}} & (1) Baseline \citep{rashkin-etal-2019-towards} & 3.41 & 3.75 & 3.72 \\
 & (3) Finetuned-EF & 3.33 & 3.67 & 3.74 \\
 & (5) DQN-Finetuned & \textbf{3.65} & \textbf{3.9} & \textbf{3.97} \\ \specialrule{0.05em}{3pt}{3pt}
\multirow{2}{*}{\textbf{Generative}} & (8) Finetuned & 3.15 & 3.52 & 3.8 \\
 & (10) PG-Prepend & \textbf{3.46} & \textbf{3.74} & \textbf{3.85} \\ \specialrule{0.05em}{3pt}{3pt}
\begin{tabular}[c]{@{}c@{}}  \textbf{Ground Truth}\end{tabular} & - & 3.51 & 3.81 & 3.88 \\ 
\bottomrule{}
\end{tabular}

\label{tab:ecf}
\end{table}

\subsection{Human Rating Results}
Results of human ratings substantiate our proposed method to explain and quantify the relationship between positivity and empathy. Table \ref{tab:ecf} shows that models finetuned by deep RL methods significantly improve all scores, especially on the empathy column. This performance meets the results in Table \ref{tab:rl} that the DQN-Finetuned retrieval model augment rewards the most. To strengthen the credibility of this result, we conduct the Pair-wise T-test. When compared the DQN-Finetuned model to other models, the p-values are all less than 0.05, which thus indicates that our DQN-Finetuned model is significantly better than others. In generative-based models, the score of the PG-EmoPrend model is relatively close to the Ground Truth. It is mainly because in Table \ref{tab:auto}, the improved Avg BLEU score of the model shows that generated sentences are similar to the Ground Truth. Moreover, the worse performance of models without finetuned by deep RL reveals that these models suffer from a deficiency in understanding the speaker's underlying emotional feelings. From the third party's viewpoints, models only finetuning on the ED dataset are far from claiming to be empathetic enough to make the speaker feel sympathy. 

Furthermore, Table \ref{tab:pair} demonstrates that humans tend to choose the DQN-Finetuned retrieval model's response as the most empathetic response. In the retrieval models, half of the participants prefer the RL-trained model to other models. Similarly, the PG-EmoPrepend model is more favored by the participants than the baseline model. In addition, the consequence that retrieval chatbot prevails the generative chatbot shows consistency with the experiments in Table \ref{tab:rl} and \ref{tab:ecf}. The results lead to concrete proof that CheerBots are more likely to produce an empathetic response through deep RL training against other chatbots. Table \ref{tab:ques_ex} is an example in the questionnaires and displays different responses according to different models. We demonstrate more examples of questionnaires and real conversations between the interlocutors and our CheerBots in Appendix \ref{sec:chatbot_ex}.

\begin{table*}
\centering
\caption{Example Responses to different kinds of Models. The setups of each model are the same as Table \ref{tab:auto}. Ground Truth is a human example sampled from the ED dataset. Given the same context, we evaluate different responses generated by various models.}
\label{tab:ques_ex}
\begin{tabular}{c m{3.5cm} p{5cm}}
\toprule{}
\textbf{Context} & \textbf{Model} & \textbf{Responses} \\
\specialrule{0.05em}{3pt}{3pt}
\multirow{18}{4cm}{\textbf{I had a job interview that I didn't really feel prepared for.}} & (1) Ret-Pretrained & Yeah, that can be nerve-racking! How did it go?\\ \specialrule{0em}{1pt}{1pt}
&\cellcolor[HTML]{EFEFEF} Human React & \cellcolor[HTML]{EFEFEF} Well, honestly I think I blew it. \\ \specialrule{0em}{1pt}{1pt}
& (3) Ret-Finetuned-EF & What about it made you feel nervous?\\ 
\specialrule{0em}{1pt}{1pt}
& \cellcolor[HTML]{EFEFEF} Human React & \cellcolor[HTML]{EFEFEF} They asked me a lot of questions that I never thought of. \\
\specialrule{0em}{1pt}{1pt}
& (5) Ret-DQN-Finetuned & Maybe it went better than you thought! If not, you still gained some experience from the day.\\ 
\specialrule{0em}{1pt}{1pt}
& \cellcolor[HTML]{EFEFEF} Human React & \cellcolor[HTML]{EFEFEF} Yeah, I think so. Thanks for the encouragement. \\
\specialrule{0em}{1pt}{1pt}
 & (8) Gen-Finetuned & Oh no. Did you get it?  \\
 \specialrule{0em}{1pt}{1pt}
 & \cellcolor[HTML]{EFEFEF} Human React & \cellcolor[HTML]{EFEFEF}Probably not, I don’t know the result yet. \\
 & (10) Gen-PG-Prepend & Oh no. I am sorry to hear that. How did it go?\\ 
  & \cellcolor[HTML]{EFEFEF} Human React & \cellcolor[HTML]{EFEFEF} I don’t think I did it very well. The question is way much harder than I thought. \\
& Ground Truth & That's rough. Did you end up getting the job?\\
 \specialrule{0em}{1pt}{1pt}
   & \cellcolor[HTML]{EFEFEF} Human React & \cellcolor[HTML]{EFEFEF} I did not get it, unfortunately.\\
\bottomrule{}
\end{tabular}
\end{table*}

\section{Conclusion}
In this paper, we introduce CheerBots, several empathetic chatbots trained by our proposed framework. Aided by the Emotion Controller, CheerBots, including retrieval-based and generative-based chatbots, can precisely retrieve or generate a response with specific emotion to interplay with the speaker, helped by the CHM. By considering changes between the speaker's current and the subsequent emotional state, we enhance the chatbots' capability to arouse empathy using deep RL. Thus, the results demonstrate CheerBots outperform the baseline chatbots and achieve SOTA. According to the performance in automatic metrics and human ratings, we firmly believe that CheerBots can chat with users fluently, coherently, and empathetically.

\section*{Broader Impact}
An empathetic chatbot might have a variety of beneficial impact across numerous applications. That is, these applications, including customer support, queries and complaints, home accompany, etc.,  bring the interlocutor more human-like, positive interactive experiences. However, some might misuse this technology deliberately.

We define a reward function that augments the positivity between interlocutors' sentences. Similarly, others could amplify the negativity between these sentences as a new reward function to make the speaker angry, sad, or anxious. This technology would provoke opposition and cause disputes among humans.

Furthermore, our model inherently runs risks of generating or retrieving biased content corresponding to the training dataset. \citet{solaiman2019release} indicates that when models learn from data collected from the Internet, sentence generation could produce offensive content corresponding to different sensitive aspects, such as gender, race, and religion.

\bibliographystyle{named}
\bibliography{references}

\medskip
\small

\clearpage
\appendix

\section{Training Details}
\label{sec:appendix}
\subsection{Training Hyperparameter}
\label{sec:hyperparameters}
Emotion Detector and Next Emotion Predictor's underlying parameters were as follows: Both Bert hidden size = 768, followed by dropout rate = 0.5, and a layer of linear projected size = 512. In Emotion Detector, we used leaky ReLU activation for the hidden layer, and the final hidden layer was followed by two branch layers for discrete and continuous labels. Their size was the number of emotion classes and Valence-Arousal, respectively. In Next Emotion Predictor, we concatenate the one-hot vector of detected emotion of the current sequence with BERT's output, then it also followed by a dropout rate was 0.5, and a layer of linear projected size was 512. Finally, we use the class prediction layer of size the number of emotion classes. Both training optimizers were Bert Adam with an initial learning rate $5\mathrm{e}{-5}$.

The configuration of the retrieval-based model is described as follows. We adopt the BERT encoder as our backbone architecture. The hidden size of the BERT encoder was 768, and its embedding size was 300. The batch size was set as 9 due to limited GPU memory resources. To consider the previous information in a training iteration, we added previous utterances into the history, and the maximum history length was up to 4 iterations. The maximum sequence length was 100 tokens. We initiate the learning rate at $1\mathrm{e}{-5}$ and used BERTADAM as our optimizer. Furthermore, when applying PG to the finetune Retrieval-based model, we trained 10 epochs. As for the DQN algorithm, we trained 20 epochs. On the other hand, the parameter of Emo-Prepending generative-based chatbot was as follows: the hidden size of the encoder output was 768. Then, the hidden vector was followed be two classification layers, the next sentence prediction of size 2 and the emotion classification layer of size the number of emotion classes. The batch size of model training was 4, and the maximum sequence length was 128 tokens. The optimizer was AdamW with an initial learning rate $6.25\mathrm{e}{-5}$. In deep RL finetuning, models were trained with a fixed batch size of 4. The number of epochs was 30.

Below is the shared configuration in DQN and PG methods. We froze the embeddings and first ten blocks of the Bert model in the Emotion Predictor to be non-trainable.  All other hyperparameters were as follows: conversation turns were 3, and the learning rate was $5\mathrm{e}{-5}$.  In models finetuned by PG, we adopted the baseline model, and we set discount = 0.99. In models finetuned by DQN, replay buffer = 5000, frequency to train online network = 4, frequency to update target network = 3000. We used a smooth L1 loss function to guide the Q-value and clipped the gradients between $\pm 1$.

\subsection{Persona Dialogue Pre-training}
\label{sec:persona}
Since the empathetic dialogue dataset \citep{rashkin-etal-2019-towards} is relatively small, finetuning only on such a dataset would limit the model's chit-chat capability. To improve the chit-chat capability of the generative-based chatbot, we firstly pre-train our generative-based model on PersonaChat \citep{zhang-etal-2018-personalizing} by following the training strategy of \citet{wolf2019transfertransfo}\footnote{https://github.com/huggingface/transfer-learning-conv-ai}. \citet{zhang-etal-2018-personalizing} established this persona-chat dataset to promote more engaging and more personal chit-chat dialogue. It includes a set of 1155 personas (character), each consisting of at least 5 profile sentences. It also contains 162,064 utterances over 10,907 dialogues, 15,602 utterances in 1000 dialogues are set aside for validation, and 15,024 utterances in 968 dialogues for test. With the pre-training process, we expect that our model can enable our model to understand how to respond according to the pre-defined persona (character).

\subsection{Infrastructure Training Details}
\label{sec:spe_details}
The description of the infrastructure used is shown in Table \ref{tab:infra}. The table also specifically introduces the number of parameters in each model.

\begin{table*}[]
\centering
\begin{tabular}{clcccc}
\toprule{}
 & Model & Params & resources & training time & train examples \\
\specialrule{0.05em}{3pt}{3pt}
\multirow{2}{*}{Emotion Controller} & Emotion Detector & 110M & 1 GPU & 0.5 hour & 22.3k \\ 
 & Emotion Predictor & same & same & 0.5 hour & 40.2k \\ 
\specialrule{0.05em}{3pt}{3pt}
Retrieval Chatbot  & Finetuned                   & 170M & 1 GPU & 0.5 day & 40.2k \\
\specialrule{0.05em}{3pt}{3pt}
\multirow{2}{*}{Generative Chatbot} & Finetuned                   & 117M & 1 GPU & 4 hour & 40.2k \\ 
                            & EmoPrepend                  & same  & same & 8.5 hour & same  \\ 
\specialrule{0.05em}{3pt}{3pt}
\multirow{4}{*}{RL finetuned} & Ret-PG-Finetuned                  & 8.1M & 1 GPU & 0.5 day & 40.2k \\
                            & Ret-DQN-Finetuned               & same & same & 2 days & same \\
                            & Gen-PG-Prepend                  & same  & same &  2 days & same \\
                            & Gen-DQN-Prepend                 & same  & same & 2.5 days & same \\

\bottomrule{}                            
\end{tabular}
\caption{Specific training details. This table demonstrates the detailed part of our training. Train examples mean the number of training utterances.}
\label{tab:infra}
\end{table*}

\begin{table}[t]
\centering
\begin{tabular}{cccc}
\toprule
\textbf{Methods} & \textbf{Top 1} & \textbf{Top 3} & \textbf{Top 5}\\
\specialrule{0.05em}{3pt}{3pt}
FastText & 43\% & - & - \\
DeepMoji (1)  & 46\% & - & - \\
DeepMoji (2)   & 46\% & - & - \\
DeepMoji (3)   & 48\% & - & - \\
\hline
\textbf{Our Discrete}  & 57\% & 80\% & 90\%\\
\textbf{Our Continuous} & 54\% & 67\% & 73\% \\
\textbf{Our All} & \textbf{60\%} & \textbf{83\%} & \textbf{90\%} \\ 
\bottomrule
\end{tabular}
\caption{Performance of emotion classification. The upper part is results of existing approaches, FastText \citep{joulin-etal-2017-bag}, DeepMoji \citep{felbo-etal-2017-using}. DeepMoji (1), (2), (3) represent its full, last, and chain-thaw methods, respectively. The lower one is the comparison among discrete, continuous, and mixed label domain. Top results are boldfaced.}
\label{tab:emo}
\end{table}

\section{Performance of Emotion Classification}
We compared our model to existing emotion classification approaches on the test set of the ED dataset with metric Accuracy (ACC), shown in Table \ref{tab:emo}. The performance of other approaches was obtained from \citet{rashkin-etal-2019-towards}. Because we reduce the ambiguity of sentences with similar emotions and incorporate emotion labels into CheerBots, our model performs better than others. In the experiment, we find out that the mixed objective functions of discrete and continuous labels can outperform either one applied individually. The reason might be that the labels are entirely independent when employing the discrete label loss, which can help the model split different categories. However, since the relationship between emotions is essential, we develop continuous label loss to determine the dependency between these emotions. In this regard, the model is capable of distinguishing each emotion corresponding to their order on the VA coordinate.

\section{Method of Conducting Human Ratings}
\label{sec:human_details}
We collect crowdsourcing evaluation on the website SurveyCake \footnote{https://www.surveycake.com/tw/}. To conduct the subjective test, we distributed our questionnaires to National Taiwan University students. Since these students are not native speakers, it is possible that some of them are not familiar with English conversation and may misunderstand the true meanings of responses listed in the questionnaire. Therefore, we simultaneously asked the participants to tell us their English conversation comprehensive ability by giving their scores in any of the three common English certification tests: TOEFL, TOEIC, and IELTS. We received a total of 249 questionnaires. According to the statistics, 87\% of participants had taken the English certifications. 6\% of the participants have taken the TOEIC test, and the average score is 830; 69\% of the participants have taken the TOEFL test, and the average score is 97; and 2\% of the participants have taken the IELTS test, and the average score is 7.0. As for the “other” option, most of them had taken the General English Proficiency Test (GEPT). This result confirms that our participants are qualified to understand the essential and underlying meanings of responses in the questionnaire.

In the ratings, participants were given various conversations. Each conversation contains three sentences: a speaker’s opening statement, a corresponding response, and a reply based on the previous response. These three sentences can be referred to $S_{input}$, $S_{res}$, and $S_{react}$. We design two tasks for participants: Independent Comparison and Pairwise Comparison. (1) For each response in the Independent Comparison, the participants were asked to score from 1 to 5 based on relevance and fluency introduced by \citep{rashkin-etal-2019-towards}. Also, they need to score the increase of the speaker's empathy level according to the whole conversation. (2) As for Pairwise Comparison, we set up several matchups that include two models' different responses according to the same $S_{input}$. Participants need to pick up a model's response that shows more empathy. Further details are shown in Appendix \ref{sec:human_details}.

In the questionnaire, participants were given various conversations. Each conversation contains three sentences: a speaker’s opening statement, a corresponding response, and a reply based on the previous response. These three sentences can be referred to $S_{input}$, $S_{res}$, and $S_{react}$. We design two tasks for participants: Independent Comparison and Pairwise Comparison. For each response in the Independent Comparison, the participants were asked to score from 1 to 5 based on relevance and fluency introduced by \citep{rashkin-etal-2019-towards}. Also, they need to score the increase of the speaker's empathy level according to the whole conversation. The indicators are defined as below:
\begin{description}
    \item [1. Empathy:] According to the whole conversation, did the response sounds sympathetic? Did the speaker feel more sympathy from the listener's response? 
    \item [2. Relevance:]  Did the response correlate to the conversation? Was it related to the topic for the conversation?
    \item [3. Fluency:] Is the responses grammatically correct? Could you realize the responses?
\end{description}
All of these conversations are rated on a Likert scale, ranging from 1 to 5: Score 1 means a firm disagreement; Score 3 means neutral; Score 5 means an undoubted approval. Participants are asked to evaluate empathy, relevance, fluency of each example before the next question. An example is shown as Table \ref{tab:ques_ex}.

As for Pairwise Comparison, we set up several matchups that include two models' different responses according to the same $S_{input}$. That is, we combine a set of particular model's response and its human reacting response and another set into a matchup. Participants need to pick up one of the two model's responses that shows more empathy between this matchup. They can choose either side of the model or choose the "Tie" option to claim these two models have equivocal impacts on the speaker.

\section{Examples of Different Chatbots}
\label{sec:chatbot_ex}
Table \ref{tab:conv1} shows a example generated by our DQN-finetuned retrieval-based chatbot. Another example is demonstrated as Table \ref{tab:conv2} generated by PG-finetuned generative-based chatbot. Also, Table \ref{tab:ques_ex} is an example in the questionnaires and displays different responses according to different models.


\begin{table*}[t]
\centering
\begin{tabular}{p{2cm}p{13cm}}
\toprule{} 
& \textbf{Conversation 1} \\
\specialrule{0.05em}{3pt}{3pt}
\textbf{Speaker} & Last week, I failed to pass the exam.\\ \specialrule{0em}{1pt}{1pt}
\rowcolor{mygray} \textbf{Our Model} & What did you fail at exactly?\\ \specialrule{0em}{1pt}{1pt}
\textbf{Speaker} & I failed at computer science. It is really hard.\\ \specialrule{0em}{1pt}{1pt}
\rowcolor{mygray}\textbf{Our Model} & That's not too bad. Computer science isn't easy by any standards.\\ \specialrule{0em}{1pt}{1pt}
\textbf{Speaker} & Thanks, I would try hard to pass the next exam.\\ \specialrule{0em}{1pt}{1pt}
\rowcolor{mygray}\textbf{Our Model} & As much as I wanted to keep my A in computer science back in freshman year, it didn't work out and I got a B.\\ \specialrule{0em}{1pt}{1pt}
\textbf{Speaker} & This tells me that computer science is really hard.\\ \specialrule{0em}{1pt}{1pt}
\rowcolor{mygray}\textbf{Our Model} & Yeah, it's a pretty tough course, but a lot of studying goes a long way.\\ \specialrule{0em}{1pt}{1pt}
\textbf{Speaker} & I would try to do my best.\\ \specialrule{0em}{1pt}{1pt}
\rowcolor{mygray}\textbf{Our Model} & I hope too. I have a couple of hard classes like Calculus II. Hope it would not crash me.\\
\bottomrule{}

\end{tabular}
\caption{Examples of Retrieval-Model Responses. To evaluate the robustness of our chatbots, we implemented an interface that allows users to chat with our dialogues system in real-time.}
\label{tab:conv1}
\end{table*}

\begin{table*}
\centering
\begin{tabular}{p{2cm}p{13cm}}
\toprule{}
 & \textbf{Conversation 2} \\
\specialrule{0.05em}{3pt}{3pt}
\textbf{Speaker} & Job interviews always make me sweat bullets, makes me uncomfortable in general to be looked at under a microscope like that.\\ \specialrule{0em}{1pt}{1pt}
\rowcolor{mygray} \textbf{Our Model} & Don’t nervous. Do you think you will get the job?\\ \specialrule{0em}{1pt}{1pt}
\textbf{Speaker} & I feel like i am prepared, but it just make me feel it's hard to stay calm.\\ \specialrule{0em}{1pt}{1pt}
\rowcolor{mygray}\textbf{Our Model} & I can understand that, but I hope you get the job.\\ \specialrule{0em}{1pt}{1pt}
\textbf{Speaker} & I know I should try to be confident with that. How can I control myself not to be nervous?\\ \specialrule{0em}{1pt}{1pt}
\rowcolor{mygray}\textbf{Our Model} & Don’t worry, just be confident and focus on the interview.\\ \specialrule{0em}{1pt}{1pt}
\textbf{Speaker} & I have told myself to do that, but when I get in a interview, I will really easily get distracted and underperformed. \\ \specialrule{0em}{1pt}{1pt}
\rowcolor{mygray}\textbf{Our Model} & yeah, I know that feeling, just focus on yourself, take a breath and you can do well.\\ \specialrule{0em}{1pt}{1pt}
\textbf{Speaker} & Thank you, I would try my best.\\ \specialrule{0em}{1pt}{1pt}
\rowcolor{mygray}\textbf{Our Model} & Good luck!\\
\bottomrule{}
\end{tabular}
\caption{Examples of Generative-Model Responses. To evaluate the robustness of our generative-based model, we implemented an interface that allows users to chat with our dialogues system in real-time.}
\label{tab:conv2}
\end{table*}

\end{document}